\documentclass[11pt,letterpaper]{article}
\usepackage{ijcnlp2017}
\usepackage{times}
\usepackage{latexsym}

\usepackage{url}
\usepackage{booktabs}
\usepackage{multirow}
\usepackage{amsmath}
\usepackage{amssymb}
\usepackage{graphicx}
\usepackage{subcaption}
\usepackage{enumitem}

\newcommand{\hl}[1]{\textbf{#1}}

\newcommand{\eg}{\textit{e.g.\,}}
\newcommand{\vs}{\textit{vs.\,}}

\newcommand{\viz}{\textit{viz.\,}}
\newcommand{\ie}{\textit{i.e.\,}}

\newcommand*{\thead}[1]{\multicolumn{1}{c}{\bfseries #1}}
\newcommand*{\tsubhead}[1]{\multicolumn{1}{c}{\textit{#1}}}

\ijcnlpfinalcopy

\def\ijcnlppaperid{1141}

\title{Utilizing Lexical Similarity between Related, Low-resource Languages for Pivot-based SMT}

\author{Anoop Kunchukuttan, Maulik Shah \\
       {\bf Pradyot Prakash}, {\bf Pushpak Bhattacharyya} \\
       Department of Computer Science and Engineering \\
       Indian Institute of Technology Bombay \\
       \texttt{\{anoopk,maulik.shah,pradyot,pb\}@cse.iitb.ac.in}
       }

\date{}

\begin{document}
\maketitle
\begin{abstract}

We investigate pivot-based translation between related languages in a low resource, phrase-based SMT setting. We show that a subword-level pivot-based SMT model using a related pivot language is substantially better than word and morpheme-level pivot models. It is also highly competitive with the best direct translation model, which is encouraging as no direct source-target training corpus is used. 
We also show that combining multiple related language pivot models can rival a direct translation model. 
Thus, \textit{the use of subwords as translation units coupled with multiple related pivot languages can compensate for the lack of a direct parallel corpus}.

\end{abstract}

\section{Introduction}
\label{sec:intro}

\textit{Related languages} are those that exhibit lexical and structural similarities on account of sharing a \textbf{common ancestry} or being in \textbf{contact for a long period of time} \cite{bhattacharyya2016statistical}. Machine Translation between related languages is a major requirement since there is substantial government, commercial and cultural communication among people speaking related languages \eg, Europe, India and South-East Asia. These constitute some of the most widely spoken languages in the world, but many of these language pairs have few or no parallel corpora. We address the scenario when no direct corpus exists between related source and target languages, but they share limited parallel corpora with a third related language.

Modelling \textbf{lexical similarity} among related languages is the key to building good-quality SMT systems with limited parallel corpora. \textit{Lexical  similarity} means that the languages share many words with similar form (spelling and pronunciation) and meaning \viz cognates, lateral borrowings or loan words from other languages \eg, {\small\texttt{blindness}} is {\small\texttt{andhapana}} in Hindi, {\small\texttt{aandhaLepaNaa}} in Marathi. 

For translation, lexical similarity can be utilized by transliteration of untranslated words while decoding \cite{durrani2010hindi} or post-processing \cite{nakov2012combining,kunchukuttan2014icon}. 
An alternative  approach involves the use of subwords as basic translation units. Subword units like character \cite{vilar2007can,tiedemann2009character}, orthographic syllables \cite{kunchukuttan2016orthographic} and byte pair encoded units \cite{kunchukuttan2017learning} have been used with varying degrees of success. 

On the other hand, if no parallel corpus is available between two languages, pivot-based SMT \cite{de2006catalanenglish,utiyama2007} provides a systematic way of using an intermediate language, called the \textit{pivot language}, to build the source-target translation system. The pivot approach makes no assumptions about source, pivot, and target language relatedness. 

Our work \textbf{brings together subword-level translation and pivot-based SMT in low resource scenarios}. We refer to orthographic syllables and byte pair encoded units as subwords. We show that using a pivot language related to both the source and target languages along with subword-level translation (i) significantly outperforms morpheme and word-level pivot translation, and (ii) is very competitive with subword-level direct translation. We also show that combining multiple pivot models using different related pivot languages can rival a direct parallel corpora trained model. To the best of our knowledge, ours is the first work that shows that \textbf{a pivot system can be very competitive with a direct system (in the restricted case of related languages)}. Previous work on morpheme and word-level pivot models with multiple pivot languages have reported lower translation scores than the direct model \cite{more2015augmenting,dabre2015leveraging}. \newcite{tiedemann2012character}'s work uses a character-level model in just one language pair of the triple (source-pivot or pivot-target) when the pivot is related to either the source or target (but not both).

\section{Proposed Solution}
\label{sec:pivot_trans}

We first train phrase-based SMT models between source-pivot (S-P) and pivot-target (P-T) language pairs using subword units, where the pivot is related to the source and target. We create a pivot translation system by combining the S-P and P-T models. If multiple pivot languages are available, linear interpolation is used to combine pivot translation models. In this section, we describe each component of our system and the design choices.

\paragraph{Subword translation units:} We explore \textit{orthographic syllable (OS)} and \textit{Byte Pair Encoded unit (BPE)} as subword units. 

The \textit{orthographic syllable}, a \textbf{linguistically motivated unit}, is a sequence of one or more consonants followed by a vowel, \ie a \textit{C$^+$V} unit (\eg \textit{spacious} would be segmented as \textit{spa ciou s}). Note that the vowel character sequence \textit{iou} represents a single vowel.

On the other hand, the \textit{BPE unit} is motivated by \textbf{statistical properties of text} and represents stable, frequent character sequences in the text (possibly linguistic units like syllables, morphemes, affixes). Given monolingual corpora, BPE units can be learnt using the Byte Pair Encoding text compression algorithm \cite{gage1994bpe}.

Both OS and BPE units are variable length units which provide appropriate context for translation between related languages. Since their vocabularies are much smaller than the morpheme and word-level models, data sparsity is not a problem. OS and BPE units have outperformed character n-gram, word and morpheme-level models for SMT between related languages \cite{kunchukuttan2016orthographic,kunchukuttan2017learning}.

While OS units are approximate syllables, BPE units are highly frequent character sequences, some of them representing different linguistic units like syllables, morphemes and affixes. While orthographic syllabification applies to writing systems which represent vowels (alphabets and abugidas), BPE can be applied to text in any writing system.

\paragraph{Training subword-level models:}
We segment the data into subwords during pre-processing and indicate word boundaries by a boundary marker (\_) as shown in the example for OS below:

{\small word}: {\scriptsize\texttt{Childhood means simplicity .}}  \\
{\small subword: {\scriptsize\texttt{ Chi ldhoo d \_ mea ns \_ si mpli ci ty \_ .}}}

For building subword-level phrase-based models, we use (a) monotonic decoding since related languages have similar word order, (b) higher order language models (10-gram) since data sparsity is a lesser concern due to small vocabulary size \cite{vilar2007can}, and (c) word-level tuning (by post-processing the decoder output during tuning) to optimize the correct translation metric \cite{nakov2012combining}. After decoding, we regenerate words from subwords (desegmentation) by concatenating subwords between consecutive occurrences of the boundary markers.

\paragraph{Pivoting using related language:} We use a language related to both the source and target language as the pivot language. We explore two widely used pivoting techniques:   phrase-table triangulation and  pipelining.

\textbf{Triangulation} \cite{utiyama2007,wu2007pivot,cohn2007machine} ``joins'' the source-pivot and pivot-target subword-level phrase-tables on the common phrases in the pivot columns, generating the pivot model's phrase-table. It recomputes the probabilities in the new source-target phrase-table, after making a few independence assumptions, as shown below: 

\begin{equation}
P(\bar{t}|\bar{s}) =  \sum\limits_{\bar{p}} P(\bar{t}|\bar{p}) P(\bar{p}|\bar{s})
\end{equation}

\noindent where, $\bar{s},\bar{p} \textrm{ and } \bar{t}$ are source, pivot and target phrases respectively.

In the \textbf{pipelining/transfer} approach \cite{utiyama2007}, a source sentence is first translated into the pivot language, and the pivot language translation is further translated into the target language using the S-P and P-T translation models respectively. To reduce cascading errors due to pipelining, we consider the top-$k$ source-pivot translations in the second stage of the pipeline (an approximation to expectation over all translation candidates). We used $k=20$ in our experiments. The translation candidates are scored as shown below:  

\begin{equation}
P(\mathbf{t}|\mathbf{s}) =  \sum\limits_{i=1}^k P(\mathbf{t}|\mathbf{p}^i)P(\mathbf{p}^i|\mathbf{s})
\end{equation}

\noindent where, $\mathbf{s},\mathbf{p}^i \textrm{ and } \mathbf{t}$ are the source, $i^{th}$ best source-pivot translation and target sentence respectively.  

\paragraph{Using Multiple Pivot Languages}: We use multiple pivot languages by combining triangulated models corresponding to different pivot languages. Linear interpolation is used  \cite{bisazza2011fill} for model combination. Interpolation weights are assigned to each phrase-table and the feature values for each phrase pair are interpolated using these weights as shown below: 

\begin{equation}
f^j(\bar{s},\bar{t}) =  \sum\limits_i \alpha_i f_i^j(\bar{s},\bar{t})
\end{equation}

$\qquad \textrm{s.t} \quad \sum\limits_i \alpha_i  =  1, \quad \alpha_i \geq 0$

\noindent where, $f^j$ is feature $j$ defined on the phrase pair $(\bar{s},\bar{t})$, $\alpha_i$ is the interpolation weight for phrase-table $i$. Phrase-table $i$ corresponds to the triangulated phrase-table using language $i$ as a pivot.

\section{Experimental Setup}
\label{sec:experiments}

\noindent\textbf{Languages:} 
We experimented with multiple languages from the two major language families of the Indian subcontinent: \textit{Indo-Aryan} branch of the Indo-European language family (Bengali, Gujarati, Hindi, Marathi, Urdu) and \textit{Dravidian} (Malayalam, Telugu, Tamil). These languages have a substantial overlap between their vocabularies due to contact over a long period \cite{emeneau1956india,subbarao2012south}.  

\noindent\textbf{Dataset:} 
We used the \textit{Indian Language Corpora Initiative (ILCI) corpus}\footnote{available on request from tdil-dc.in} for our experiments \cite{jha2012ilci}.
The data split is as follows -- \textbf{training: 44,777, tuning: 1K, test: 2K} sentences.
Language models for word-level systems were trained on the target side of training corpora  plus monolingual corpora from various sources 
[hin: 10M \cite{hindencorp:lrec:2014}, urd: 5M \cite{jawaid2014urdu}, tam: 1M \cite{ramasamy-bojar-zabokrtsky:2012:MTPIL}, mar: 1.8M (news websites), mal: 200K, ben: 400K, pan: 100K, guj:400K, tel: 600K \cite{quasthoff2006corpus} sentences]. 
We used the target side of parallel corpora for morpheme, OS, BPE and character-level LMs. 

\noindent\textbf{System details:} 
We trained PBSMT systems for all translation units using \textit{Moses} \cite{koehn2007moses} with \textit{grow-diag-final-and} heuristic for symmetrization of alignments, and Batch MIRA \cite{cherry2012batch} for tuning. Subword-level representation of sentences is long, hence we speed up decoding by using cube pruning with a smaller beam size (pop-limit=1000) for OS and BPE-level models. This setting has been shown to have minimal impact on translation quality \cite{kunchukuttan2016faster}. 

We trained 5-gram LMs with Kneser-Ney smoothing for word and morpheme-level models, and 10-gram LMs for OS, BPE, character-level models. We used the \textit{Indic NLP library}\footnote{{\tiny \url{http://anoopkunchukuttan.github.io/indic_nlp_library}}}  for orthographic syllabification, the \textit{subword-nmt library}\footnote{{\tiny \url{https://github.com/rsennrich/subword-nmt}}} for training BPE models and \textit{Morfessor} \cite{virpioja2013morfessor} for morphological segmentation. These unsupervised morphological analyzers for Indian languages, described in \newcite{kunchukuttan2014icon}, are trained on the ILCI corpus and the Leipzig corpus \cite{quasthoff2006corpus}. The BPE vocabulary size was chosen to match OS vocab size. We use \textit{tmtriangulate}\footnote{{\tiny \url{github.com/tamhd/MultiMT}}} for phrase-table triangulation and  \textit{combine-ptables} \cite{bisazza2011fill} for linear interpolation of phrase-tables. 

\noindent\textbf{Evaluation:} The primary evaluation metric is word-level BLEU \cite{papineni2002bleu}. We also report LeBLEU \cite{virpioja2015lebleu} scores in the appendix. LeBLEU is a variant of BLEU that does soft-matching of words and has been shown to be better for morphologically rich languages. We use bootstrap resampling for testing statistical significance \cite{koehn2004statistical}. 

\begin{table}
\setlength{\tabcolsep}{3pt}
\centering
{\footnotesize
\begin{tabular}{lrrrrrr}

\toprule
 \thead{Lang Triple} & \thead{Word} & \thead{Morph} & \thead{BPE} & \thead{OS} & \thead{Char} \\
\midrule
mar-guj-hin  & 30.23 & 36.49 & 39.05 & \hl{39.81}$^\dagger$ & 34.32\\
mar-hin-ben  & 16.63 & 21.04 & 22.46 & \hl{22.92}$^\dagger$ & 17.00 \\ 
mal-tel-tam  & 4.55 & 6.19 & \hl{7.69}$^\dagger$ & 7.19 & 3.51 \\ 
tel-mal-tam  & 5.13 & 8.29 & \hl{9.84}$^\dagger$ & 8.39 & 4.26 \\ 
hin-tel-mal  & 5.29 & 8.32 & 9.57 & \hl{9.67} & 6.24 \\ 
mal-tel-hin  & 10.03 & 13.06 & \hl{17.68} & 17.26 & 9.12 \\ 
\midrule
mal-urd-hin & 7.70 & 11.29 & \hl{16.40} & NA & 7.46\\ 
urd-hin-mal & 5.58 &  6.64 &  \hl{7.58} & NA & 4.07 \\ 
\midrule
{\tiny \textit{average \% change}}  & \multirow{2}{*}{\tiny\textit{(+66,+57)\%}} & \multirow{2}{*}{\tiny\textit{(+21,+14)\%}} &  &  &  \multirow{2}{*}{\tiny\textit{(+81,+66)\%}} \\
{\tiny \textit{w.r.t (BPE,OS)}} &  &  &  &  &   \\
\bottomrule
\end{tabular}
}
\caption{{\small Comparison of triangulation for various translation units (BLEU). Lang triple refers to the source-pivot-target languages. Scores in \textbf{bold} indicate highest values for the language triple. $\dagger$ means difference between OS and BPE scores is statistically significant ($p<0.05$). NA: OS segmentations cannot be done for Urdu. The last row shows average change in BLEU scores for word, morpheme and character-level model compared to the OS and BPE-level models.}}
\label{tbl:compare_subwords}
\end{table}

\section{Results and Discussion}
\label{sec:res_subword}

In this section, we discuss and analyze the results of our experiments.

\subsection{Comparison of Different Subword Units}

Table \ref{tbl:compare_subwords} compares pivot-based SMT systems built with different units. We observe that \textit{the OS and BPE-level pivot models significantly outperform word, morpheme and character-level pivot models} (average improvements above 55\% over word-level and 14\% over morpheme-level). The greatest improvement is observed when the source and target languages belong to different families (though they have a contact relationship), showing that subword-level models can utilize the lexical similarity between languages. Translation between agglutinative Dravidian languages also shows a major improvement. The OS and BPE models are comparable in performance. However, unlike OS, BPE segmentation can also be applied to translations involving languages with non-alphabetic scripts (like Urdu) and show significant improvement in those cases also. Evaluation with LeBLEU \cite{virpioja2015lebleu}, a metric suited for morphologically rich languages, shows similar trends (results in Appendix \ref{apx:lebleu}). For brevity, we report BLEU scores in subsequent experiments.

\begin{table}
\centering
{\footnotesize
\begin{tabular}{lrrrr}
\toprule
\multirow{2}{*}{\textbf{Lang Triple}} & \multicolumn{2}{c}{\textbf{BPE}}       & \multicolumn{2}{c}{\textbf{OS}}   \\ 
\cmidrule(lr){2-3}
\cmidrule(lr){4-5}
			       & \tsubhead{pip}  	& \tsubhead{tri}        & \tsubhead{pip}  	& \tsubhead{tri} \\ 
\midrule
mar-guj-hin & 38.25 & \hl{39.05}$^\dagger$ & 38.11 & \hl{39.81}$^\dagger$ \\
mar-hin-ben & \hl{22.50} & 22.46$\ $ & 22.83 & \hl{22.92}$\ $ \\
mal-tel-tam & \hl{7.84}  & 7.69$\ $  & 6.94  & \hl{7.19}$\ $  \\
tel-mal-tam & 8.47  & \hl{9.84}$^\dagger$  & 7.96  & \hl{8.39}$^\dagger$  \\
hin-tel-mal & 9.31  & \hl{9.57}$\ $  & 9.31  & \hl{9.67}$^\dagger$  \\
mal-tel-hin & 17.39 & \hl{17.68}$\ $ & 16.96 & \hl{17.26}$\ $ \\  
\midrule
mal-urd-hin & \hl{16.93}$^\dagger$ & 16.40 & NA & NA$\ $ \\  
urd-hin-mal & \hl{8.83}$^\dagger$ &  7.58 & NA & NA$\ $ \\  
\bottomrule
\end{tabular}
}
\caption{{\small Comparison of pipelining (\textit{pip}) and triangulation (\textit{tri}) approaches for OS and BPE (BLEU). $\dagger$ means difference between \textit{pip} and \textit{tri} is statistically significant ($p<0.05$)}}
\label{tbl:compare_triangulation_transfer}
\end{table}

Subword-level models outperform other units for the pipelining approach to pivoting too. Triangulation and pipelining approaches are comparable for BPE and OS models (See Table \ref{tbl:compare_triangulation_transfer}). Hence, we report results for only the triangulation approach in subsequent experiments.

\subsection{Why is Subword-level Pivot SMT better?}

\begin{table}[t]
\centering
{\small
\begin{tabular}{lrrrrr}
\toprule
\thead{Lang Triple} &  \thead{Word} &  \thead{Morph} &  \thead{BPE} & \thead{OS} & \thead{Char} \\
\midrule
mar-guj-hin   & 0.64 & 1.39 & 1.74 & 2.33 & 3.04 \\ 
mar-hin-ben   & 0.58 & 1.36 & 1.71 & 2.6 & 3.47 \\ 
mal-tel-tam   & 0.61 & 2.32 & 3.27 & 4.19 & 2.58 \\ 
tel-mal-tam   & 0.75 & 2.82 & 4.09 & 2.76 & 2.42 \\ 
hin-tel-mal   & 0.56 & 2.08 & 2.86 & 2.97 & 2.25 \\ 
mal-tel-hin   & 0.55 & 2.28 & 2.85 & 3.56 & 2.57 \\ 
\midrule
mal-urd-hin   & 0.25 & 1.16 & 1.84 & NA & 2.05 \\ 
urd-hin-mal   & 0.42 & 0.79 & 1.62 & NA & 1.47 \\ 
\bottomrule
\end{tabular}
}
\caption{Ratio of triangulated to component phrase-table sizes. We use the size of larger of the component phrase-tables to compute the ratio.}
\label{tbl:pivot_ratio}
\end{table}

Subword-level pivot models are better than other units for two reasons. One, \textit{the underlying S-P and P-T translation models are better} (\eg 16\% and 3\% average improvement over word and morpheme-level models for OS). 
Two, the triangulation process involves an \textit{inner join} on pivot language phrases common to the S-P and P-T phrase-tables. This causes data sparsity issues due to the large word and morpheme phrase-table vocabulary \cite{dabre2015leveraging,more2015augmenting}. On the other hand, \textit{the OS and BPE phrase-table vocabularies are smaller, so the impact of sparsity is limited}. This effect can be observed by comparing the ratio of the triangulated phrase-table (S-P-T) with the component phrase-tables (S-P and P-T). The size of the triangulated phrase-table is less than the size of the underlying tables at the word-level, while it increases by a few multiples for subword-level models (see Table \ref{tbl:pivot_ratio}). 

\subsection{Comparison of Pivot \& Direct Models}

\begin{table}
\setlength{\tabcolsep}{3pt}
\centering
{\footnotesize
\begin{tabular}{lrrrrrr}
\toprule
\multirow{2}{*}{\textbf{Lang Triple}} & \thead{Pivot} & \multicolumn{4}{c}{\textbf{Direct}} & \thead{Pivot}\\
\cmidrule(lr){2-2}
\cmidrule(lr){3-6}
\cmidrule(lr){7-7}
  & \thead{BPE}  & \thead{BPE} & \thead{Word} & \thead{Morph} &  \thead{OS} &\thead{OS} \\
\midrule
mar-guj-hin  & 39.05 & 43.19 & 38.87 & 42.81 & 43.69 & 39.81 \\
mar-hin-ben  & 22.46 & 24.13 & 21.13 & 23.96 & 23.53 & 22.92 \\
mal-tel-tam  &  7.69 &  8.67 &  6.38 &  7.61 &  7.84 &  7.19 \\
tel-mal-tam  &  9.84 & 11.61 &  9.58 & 10.61 & 10.52 &  8.39 \\
hin-tel-mal  &  9.57 & 10.73 &  8.55 &  9.23 & 10.46 &  9.67 \\
mal-tel-hin  & 17.68 & 20.54 & 15.18 & 17.08 & 18.44 & 17.26 \\
\midrule
mal-urd-hin  &  16.4 & 20.54 & 15.18 & 17.08 & 18.44 &  NA   \\  
urd-hin-mal  &  7.58 &  8.44 &  6.49 &  7.05 &  NA   &  NA   \\ 
\bottomrule
\end{tabular}
}
\caption{Pivot \vs Direct translation (BLEU)}
\label{tbl:compare_with_direct}
\end{table}

We compared the OS and BPE-level models with direct models trained on different translation units (see Table \ref{tbl:compare_with_direct}). 
These subword-level pivot models outperform word-level direct models by 5-10\%, which is encouraging. Remarkably, the subword-level pivot model is competitive with the morpheme-level models (about 95\% of the morpheme BLEU score).  The subword-level pivot models are competitive with the best performing direct counterparts too (about 90\% of the direct system BLEU score).  To put this fact in perspective, the BLEU scores of morpheme and word-level pivot systems are far below their corresponding direct systems (about 15\% and 35\% respectively). \textit{These observations strongly suggest that pivoting at the subword-level can better reconstruct the direct translation system than word and morpheme-level pivot systems.}

\subsection{Multiple Pivot Languages}
\label{sec:res_multilinguality}

\begin{table}
\centering
{\footnotesize
\begin{tabular}{lllll}
\toprule
\multirow{2}{*}{\textbf{Model}} &  \multicolumn{2}{c}{\textbf{mar-ben}} & \multicolumn{2}{c}{\textbf{mal-hin}} \\
\cmidrule(lr){2-3}
\cmidrule(lr){4-5}
 & \thead{OS} & \thead{BPE} & \thead{OS} & \thead{BPE}\\
\midrule
\multirow{2}{*}{best pivot} & 22.92  & 22.46 & 17.52 & 18.47 \\
           & (\textit{hin})  & (\textit{hin})  & (\textit{tel})   & (\textit{guj}) \\
direct            & 23.53   & 24.13 & 18.44  & 20.54 \\
all pivots        & 23.69   & 23.20$^\dagger$ & 19.12$^\dagger$  &  20.28 \\
direct+all pivots &  24.41$^\ddagger$  & \hl{24.49}$^\ddagger$ & 19.44$^\ddagger$  & \hl{20.93}$^\ddagger$ \\
\bottomrule
\end{tabular}
}
\caption{{\small Combination of multiple pivots (BLEU). \textbf{Pivots used for} (i) mar-ben: guj, hin, pan (ii) mal-hin: tel, mar, guj. Best pivot language indicated in brackets. Statistically significant difference from \textit{direct} is indicated for: \textit{all pivots}($\dagger$) and \textit{direct+all pivots}($\ddagger$) ($p<0.05$).}}
\label{tbl:multiple_pivots}
\end{table}

We investigated if combining multiple pivot translation models can be a substitute for the direct translation model. \textit{Direct model} refers to translation system built using the source-target parallel corpus. Using linear interpolation with \textit{equal weights}, we combined pivot translation models trained on different pivot languages. Table \ref{tbl:multiple_pivots} shows that \textit{the combination of multiple pivot language models outperformed the individual pivot models, and is comparable to the direct translation system}. Previous studies have shown that word and morpheme-level multiple pivot systems were not competitive with the direct system, possibly due to the effect of sparsity on triangulation \cite{more2015augmenting,dabre2015leveraging}. Our results show that once the ill-effects of data sparsity are reduced due to the use of subword models, multiple pivot languages can maximize translation performance because: (i) they bring in more translation options, and (ii) they improve the estimates of feature values with evidence from multiple languages. Linear interpolation of the direct system with all the pivot systems with equal interpolation weights also benefitted the translation system.  Thus, \textit{multilinguality helps overcome the lack of parallel corpora between the two languages}. 

\subsection{Cross-Domain Translation}

We also investigated if the OS and BPE-level pivot models are robust to domain change by evaluating the pivot and direct translation models trained on tourism and health domains on an agriculture domain test set of 1000 sentences (results in Table \ref{tbl:cross_domain}). For cross-domain translation too, the subword-level pivot models outperform morpheme-level pivot models and are comparable to a direct morpheme-level model. The OS and BPE-level models systems experience much lesser drop in BLEU scores \textit{vis-a-vis} direct models, in contrast to the  morpheme-level models. 
Since morpheme-level pivot models encounter unknown vocabulary in a new domain, they are less resistant to domain change than subword-level models. 

\section{Conclusion and Future Work}
\label{sec:conclusion}

\begin{table}
\setlength{\tabcolsep}{3pt}
\centering
{\footnotesize
\begin{tabular}{lrrrrrrrr}
\toprule
\multirow{2}{*}{\textbf{Lang Triple}} & \multicolumn{3}{c}{\textbf{Pivot}} & \multicolumn{3}{c}{\textbf{Direct}} \\
\cmidrule(lr){2-4}
\cmidrule(lr){5-7}
 & \thead{Morph} & \thead{OS} & \thead{BPE} & \thead{Morph} & \thead{OS} & \thead{BPE} \\
\midrule
hin-tel-mal & 4.72 &  5.96  &  6.00 &  5.99 &   6.26 &  6.37 \\
mal-tel-hin & 8.29 & 11.33  & 10.94 & 11.12 &  13.32 & 14.45 \\
mal-tel-tam & 4.41 &  5.82  &  5.85 &  5.84 &   5.88 &  6.75 \\
\bottomrule
\end{tabular}
}
\caption{Cross domain translation (BLEU)}
\label{tbl:cross_domain}
\end{table}

We show that pivot translation between related languages can be competitive with direct translation if a \textit{related pivot language} is used and \textit{subword units} are used to represent the data. Subword units make pivot models competitive  by (i) utilizing lexical similarity to improve the underlying S-P and P-T translation models, and (ii) reducing losses in pivoting (owing to  small vocabulary). Combining multiple related pivot models can further improve translation. Our SMT pivot translation work is useful for low resource settings, while current NMT systems require large-scale resources for good performance. We plan to explore  multilingual NMT in conjunction with subword representation between related languages with a focus on reducing corpus requirements. Currently, these ideas are being actively explored in the research community in a general setting. 

\bibliography{ijcnlp2017_short_pivot_subword}
\bibliographystyle{ijcnlp2017}

\appendix

\section{LeBLEU Scores}
\label{apx:lebleu}

Table \ref{tbl:lebleu_pivot_subword} shows LeBLEU scores for the experiments using phrase-triangulation. We observe that the same trends hold as with BLEU scores.

\makeatletter
\setlength{\@fptop}{0pt}
\makeatother

\begin{table}[tp!]

\begin{subtable}{\columnwidth}
\setlength{\tabcolsep}{3pt}
\centering
{\footnotesize
\begin{tabular}{lrrrrrr}
\toprule
\thead{Lang Triple}  & \thead{Word} & \thead{Morph} & \thead{BPE} & \thead{OS} & \thead{Char} \\
\midrule
mar-guj-hin   & 0.692 & 0.725 & 0.737 & \hl{0.747} & 0.713 \\
mar-hin-ben   & 0.505 & 0.616 & 0.638 & \hl{0.646} & 0.577 \\
mal-tel-tam   & 0.247 & 0.364 & \hl{0.426} & 0.407 & 0.213 \\
tel-mal-tam   & 0.242 & 0.433 & \hl{0.485} & 0.441 & 0.392 \\
hin-tel-mal   & 0.291 & 0.376 & 0.420 & \hl{0.432} & 0.306 \\
mal-tel-hin   & 0.247 & 0.364 & \hl{0.426} & 0.404 & 0.213 \\
\midrule
mal-urd-hin   & 0.328 & 0.436 & \hl{0.501} & NA & 0.377 \\
urd-hin-mal   & 0.313 & 0.353 & \hl{0.420} & NA & 0.323 \\
\midrule
{\tiny \textit{average \% change}}  & \multirow{2}{*}{\tiny\textit{(+51,+49)\%}} & \multirow{2}{*}{\tiny\textit{(+12,+8)\%}} &  &  &  \multirow{2}{*}{\tiny\textit{(+42,+42)\%}}\\
{\tiny \textit{w.r.t (BPE,OS)}} &  &  &  &  &   \\
\bottomrule
\end{tabular}
}
\caption{Comparison of phrase-triangulation for various subwords}
\label{tbl:compare_subwords_lebleu}
\end{subtable}

\vspace{0.25cm}

\begin{subtable}{\columnwidth}
\setlength{\tabcolsep}{3pt}
\centering
{\footnotesize
\begin{tabular}{lrrrrrr}
\toprule
\multirow{2}{*}{\textbf{Lang Triple}} & \thead{Pivot} & \multicolumn{4}{c}{\textbf{Direct}} & \thead{Pivot}\\
\cmidrule(lr){2-2}
\cmidrule(lr){3-6}
\cmidrule(lr){7-7}
 & \thead{BPE}  & \thead{BPE} & \thead{Word} & \thead{Morph} &  \thead{OS} &\thead{OS} \\
\midrule
mar-guj-hin   & 0.737 & 0.766 & 0.746 & 0.767 & 0.766 & 0.747 \\
mar-hin-ben   & 0.638 & 0.653 & 0.568 & 0.645 & 0.656 & 0.646 \\
mal-tel-tam   & 0.426 & 0.465 & 0.314 & 0.409 & 0.447 & 0.407 \\
tel-mal-tam   & 0.485 & 0.530 & 0.410 & 0.511 & 0.534 & 0.441 \\
hin-tel-mal   & 0.420 & 0.468 & 0.393 & 0.436 & 0.477 & 0.432 \\
mal-tel-hin   & 0.426 & 0.565 & 0.460 & 0.528 & 0.551 & 0.404 \\
\midrule
mal-urd-hin   & 0.501 & 0.565 & 0.460 & 0.528 & 0.551 & NA \\
urd-hin-mal   & 0.420 & 0.416 & 0.350 & 0.379 & NA & NA \\
\bottomrule
\end{tabular}
}
\caption{Pivot \vs direct translation}
\label{tbl:compare_with_direct_lebleu}
\end{subtable}

\caption{LeBLEU Scores}
\label{tbl:lebleu_pivot_subword}
\end{table}

\end{document}